\pdfoutput=1
\documentclass[11pt]{article}

\usepackage[english]{babel}

\usepackage[utf8]{inputenc}

\usepackage{csquotes}
\usepackage{color}
\usepackage{amsmath}
\usepackage{balance}
\usepackage[linesnumbered,ruled]{algorithm2e}
\usepackage{multirow}
\usepackage{makecell}
\usepackage{booktabs}

\usepackage[final]{acl}
\usepackage{times}
\usepackage{latexsym}

\usepackage[T1]{fontenc}
\usepackage{microtype}

\usepackage{inconsolata}

\usepackage{graphicx}
\usepackage{changepage}

\title{Retrieval Augmented Spelling Correction for E-Commerce Applications}

\author{Xuan Guo \\
  Amazon.com Inc\\
  \texttt{xuangu@amazon.com} \\
  \And
  Rohit Patki \\
  Amazon.com Inc\\
  \texttt{patkir@amazon.com} \\
  \AND
  Dante Everaert \\
  Amazon.com Inc\\
  \texttt{danteev@amazon.com} \\
  \And
  Christopher Potts \\
  Stanford University \\
  \texttt{cgpotts@stanford.edu} \\}

\begin{document}
\maketitle
\begin{abstract}
The rapid introduction of new brand names into everyday language poses a unique challenge for e-commerce spelling correction services, which must distinguish genuine misspellings from novel brand names that use unconventional spelling. We seek to address this challenge via Retrieval Augmented Generation (RAG). On this approach, product names are retrieved from a catalog and incorporated into the context used by a large language model (LLM) that has been fine-tuned to do contextual spelling correction. Through quantitative evaluation and qualitative error analyses, we find improvements in spelling correction utilizing the RAG framework beyond a stand-alone LLM. We also demonstrate the value of additional finetuning of the LLM to incorporate retrieved context.
\end{abstract}

\section{Introduction}\label{sec:intro}

New brand names are continuously introduced, and many of them use unconventional spelling to create specific associations while still ensuring that the brand is unique and memorable. Prominent examples include ``playgro'' vs.\ ``playground'', ``biotanicals'' vs.\ ``botanicals'', and ``hygeeni'' vs.\ ``hygiene''. Such cases pose a significant challenge for e-commerce spelling correction services, which are prone to over-correcting such terms, especially completely novel ones.

In this paper, we seek to address this challenge by leveraging Retrieval Augmented Generation (RAG). On this approach, the user's query is passed to a retrieval module that seeks to find relevant items from a product catalog. The retrieved items are then incorporated into a prompt to a large language model (LLM) that predicts a correct spelling for the user's query.

We report on a wide range of experiments with different retrieval models and LLMs. We find that the RAG-based approach consistently leads to large performance improvements with only minor latency increases. In addition, we explore methods for fine-tuning the LLM to make better use of retrieved contexts, and we find that this leads to very substantial improvements, with the largest gains coming from queries that contain brand names.

\section{Related Work}\label{sec:related_work}

\paragraph{Retrieval Augmentation}
Retrieval has proven highly effective across a wide range of knowledge intensive tasks. Early works such as DrQA~\cite{chen2017reading} and Dense Passage Retrieval (DPR; \citealt{karpukhin2020dense}) laid important groundwork by integrating retrieval with neural models to enhance question-answering accuracy and facilitate knowledge access. REALM~\cite{guu2020retrieval} introduced unsupervised retrieval for language model pre-training, and~\citet{lewis2020retrieval} combined retrieval with generation to tackle knowledge intensive tasks. RETRO~\cite{borgeaud2022improving} scaled these ideas by accessing a vast database of tokens for contextual relevance across large datasets. \citet{khattab-etal-2021-relevance} weakly supervise the ColBERT neural retrieval model to improve performance on open-domain question answering. Overall, these approaches seem to help systems provide up-to-date knowledge and reduce hallucinations \cite{retrieval-lm-tutorial}. Despite these advances, the specific challenges of contextual spelling correction in e-commerce settings, particularly for dynamically evolving brand names, remain underexplored.

\paragraph{Retrieval Augmented Fine-Tuning}
Retrieval Augmented Fine-Tuning (RAFT) adapts models by fine-tuning them for specific tasks like question answering, with a strong focus on handling irrelevant documents to boost accuracy~\cite{zhang2024raft}. Similarly, Atlas~\cite{izacard2023atlas} demonstrates the effectiveness of retrieval-augmented models for few-shot learning by integrating retrieved content during both pre-training and fine-tuning phases. While other models like RPT~\cite{rubin2023long} and REPLUG~\cite{shi2023replug} also combine retrieval with generative capabilities, they either focus on black-box integration or training from scratch without fine-tuning on external context. Our approach, akin to RAFT, fine-tunes the LLM with task-specific data, distinguishing between relevant and irrelevant contexts, especially for complex non-standard brand names. This underscores the value of targeted retrieval in specialized tasks like e-commerce spelling correction.

\paragraph{Contextualized Spelling Correction}
Related studies in character-level and multilingual spelling correction highlight the importance of context and fine-grained adjustments. For example, \citet{zhang2023multi} show that using multiple teacher models improves spelling correction across diverse languages, underscoring the value of context-specific training. \citet{huang2022inducing} introduce structural interventions at the subword level to improve spelling correction, particularly for morphologically complex terms. Unlike these methods, which are tailored to specific contexts, our RAG-based approach generalizes to evolving and unconventional brand names by retrieving up-to-date context on demand, enhancing adaptability and robustness in real-world e-commerce applications.

Contextualized spell-checking methods~\cite{song2023contextual,wang2023improving} emphasize the value of integrating external knowledge to handle domain-specific terms, demonstrating the benefits of user-specific data for improved spelling accuracy. Unlike these methods, which use prompt conditioning and attention mechanisms to incorporate context, our approach employs RAG to meet the unique demands of an evolving e-commerce catalog. By leveraging RAG, we dynamically integrate new and modified brand names directly from the catalog, enabling adaptation to non-standard lexicons unique to brand names. This retrieval-based method helps address the continuously updated nature of brand catalogs in ways that prompt-tuning or attention-based techniques alone may not fully resolve, particularly for ambiguous or unconventional spellings.

\section{Approach}\label{sec:method}

We now present our approach to spelling correction, which leverages Retrieval Augmented Generation (RAG) and optionally includes a phase of fine-tuning the LLM to make better use of context.

\subsection{Language Models}\label{sec:llms} 

We experiment primarily with \mbox{Mistral-7B}~\cite{jiang2023mistral} (specifically, \texttt{open-mistral-7b} v0.1) and Claude-3-sonnet (\texttt{claude-3-sonnet-20240229} v1:0). We chose Mistral-7B because it is highly effective for its size (see Section~\ref{sec:zero_shot_llm}), and we chose Claude-3-sonnet as a representative of a much larger class of LLMs. We would expect to see similar results for other LLMs, even larger and more capable ones, because our central challenge is making accurate predictions about novel brand names.

\subsection{Retrieval Models}\label{sec:retrievers}

We evaluate three main retrieval methods:
\begin{enumerate}
\item \textbf{BM25} \citep{robertson2009probabilistic} is a traditional, time-tested n-gram-based retrieval model. We expect it to excel where exact matching suffices but may struggle where fuzzy or semantic matching is called for.
\item \textbf{Fuzzy BM25} combines traditional BM25 with fuzzy matching. This allows for minor spelling errors that are common in e-commerce queries. For instance, it can match ``air fryer cusinart'' to both ``air fryer'' and ``cuisinart''.
\item \textbf{ColBERT} \citep{khattab2020colbert,Santhanam:Khattab-etal:2022,santhanam2021colbertv2} is a neural retrieval model that represents queries and documents with token-level vectors. We expect this model to excel at retrieving semantically relevant terms.
\end{enumerate}
In all cases, we index our product catalog. For ColBERT, this is done using a pretrained ColBERTv2 model checkpoint released by the ColBERT team.\footnote{\href{https://huggingface.co/colbert-ir/colbertv2.0}{huggingface.co/colbert-ir/colbertv2.0}}

\subsection{Prompt Design}\label{sec:LLM_component}

The following is the LLM prompt template that we use where the retrieved items are provided as context:

{\small
\begin{verbatim}
### Instruction:
Provide spelling correction for given query 
if necessary, referring to the provided context
if it's relevant.
### Context:
{context}
### Query:
{input}
### Correction:
\end{verbatim}
}

\noindent
Retrieved items are formatted as a comma-separated list and placed in the \texttt{context} field. We use the top 3 or 4 retrieved items, with the precise number controlled by the maximum prompt length we permit in our fine-tuning runs. (Future work could explore the effects of including more context items at inference time.)

\subsection{LLM Fine-Tuning}\label{sec:finetune_data_elicit}

Our approach includes an optional step of fine-tuning the LLM to make more effective use of context. We consider two variants.

For \textbf{Basic Fine-Tuning}, the training dataset consists of 50K \texttt{<input query, label query>} pairs derived from search logs to create a training dataset. Here, \texttt{label query} reflects user-validated corrections. For this fine-tuning, we remove the ``referring to the provided context if it's relevant'' wording from \texttt{Instruction} of the prompt in Section~\ref{sec:LLM_component}, with the \texttt{context} field also removed and the desired output appended to the end followed by \texttt{\#\#\#} on a newline.

For \textbf{Contextual Fine-Tuning}, we begin with the same dataset of 50K pairs, but now we retrieve context for each \texttt{<input query>}, forming \texttt{<input query, context, label query>} triples. This added context, derived through the same retrieval mechanism used in RAG, helps teach the model how to make use of the \texttt{context} field. The prompt has the same format as the one in Section~\ref{sec:LLM_component} but with the desired output appended to the end followed by \texttt{\#\#\#} on a newline.

For both variants, the fine-tuning process is simply additional language model training using strings formatted from our prompt templates. Further evaluation details, including metrics, are covered in Section~\ref{sec:experiment}.

\begin{table}[tp]
\centering
  \begin{tabular}{l l r}
    \toprule
    LLM (no finetune) & Size & F1 \\
    \midrule
    Flan UL2 & 20B & 13.0\\
    mT0 & 13B & 19.3\\
    Flan T5 & 11B & 20.0\\
    mGPT & 13B & 21.7\\
    Mistral & \phantom{0}7B & 28.1\\
    Claude-3-sonnet & 70B$^{\ast}$ & 34.7 \\
    Mixtral & 47B & 57.4\\
  \bottomrule
\end{tabular}
\caption{Results for zero-shot LLMs. The largest models achieve the best results, and Mistral-7B achieves excellent results for its size. ${}^{\ast}$The size given for Claude-3-sonnet is a guess based on the comparative performance of the model relative to others of known size.}
\label{tab:generative_performance}
\end{table}

\section{Experiments}\label{sec:experiment}

\subsection{Evaluation Data}\label{sec:data}

Our evaluation dataset is sourced from search logs collected between 2021 and 2023. Each data point consists of an \texttt{<input query, label query>} pair. The \texttt{input query} refers to the user's original search query, while the \texttt{label query} is obtained from annotators. We conducted stratified sampling to arrive at a 10K \texttt{input query} set, which was designed to promote the diversity of the query population, particularly with regard to the presence of misspellings (roughly 1/4) and brand names (roughly 1/3 cases with a brand name). To ensure label quality, we applied a ``2+1'' annotation method. That is, two annotators initially labeled each query; if they disagreed, an auditor made the final determination. This process included specific instructions to guide annotators on the e-commerce context. We use this dataset to evaluate all experiments in this paper.

\subsection{Metrics}

Our primary metric for evaluating spelling correction quality is the F1 score, which is the harmonic mean of precision and recall. Precision reflects the proportion of model-predicted corrections that match the gold standard annotations, while recall measures the proportion of required corrections that the model identifies correctly. We rely on exact match criteria, where two strings are considered equal after punctuation removal. All F1 scores are reported as percentages for clarity.

\subsection{Zero-Shot LLM Performance}\label{sec:zero_shot_llm}

Table~\ref{tab:generative_performance} reports baseline results for a range of different LLMs used without any retrieval. The prompt template used for this is the same one as in Section~\ref{sec:LLM_component} without the \texttt{context} field and its mentions in the \texttt{Instruction} section. The top-performing model by a large margin is Mixtral-47B, followed by Claude-3-sonnet ($\approx$70B, estimated). The much smaller Mistral-7B model~\cite{jiang2023mistral} is reasonably competitive, though, and it represents a better balance of costs and performance, especially for high-volume services like spelling correction. 

\subsection{RAG with a Frozen LLM}
\label{sec:rag}
\begin{table}[tp]
  \resizebox{1.0\linewidth}{!}{
  \setlength{\tabcolsep}{4pt}
  \begin{tabular}{@{} l l r r @{}}
    \toprule
    Retriever & LM & Doc index & F1 \\
    \midrule
    BM25 & Mistral-7B & 572K & 25.4\\
    Fuzzy BM25 & Mistral-7B & 60K & 31.7\\
    Fuzzy BM25 & Mistral-7B & 572K & 34.6\\
    ColBERT & Mistral-7B  & 60K & 35.8\\
    ColBERT & Mistral-7B & 572K & 35.9\\
    \midrule
    BM25 & Claude-3-sonnet & 572K & 26.2\\
    Fuzzy BM25 & Claude-3-sonnet & 60K & 30.4\\
    Fuzzy BM25 & Claude-3-sonnet & 572K & 34.8\\
    ColBERT & Claude-3-sonnet & 60K & 29.8\\
    ColBERT & Claude-3-sonnet & 572K & 39.3\\
  \bottomrule
  \end{tabular}
  }
  \caption{Results for RAG with frozen LLMs. The best configurations use ColBERT and a larger document index. Both LLMs are substantially improved by RAG.}
  \label{tab:retriever_performance}
\end{table}

\begin{table*}[tp]
  \resizebox{1\linewidth}{!}{
  \begin{tabular}{p{0.3cm} p{2.5cm} p{6.7cm} p{2.5cm} p{2.5cm}}
    \toprule
    \# & User query & Retrieved items & RAG generation & Ground truth \\
    \midrule
    1 & \colorbox{red!30}{sumbrella} umbrella outdoor patio & abba patio umbrella, patio umbrella \colorbox{green!30}{sunbrella}, \colorbox{green!30}{sunbrella} umbrellas, hampton bay patio umbrella & \colorbox{green!30}{sunbrella} umbrella outdoor patio & \colorbox{green!30}{sunbrella} umbrella outdoor patio\\
    2 & \colorbox{red!30}{niwax} ne tx direct & \colorbox{green!30}{nikwax} tx direct,tx direct spray on,tx direct & \colorbox{green!30}{nikwax} ne tx direct & \colorbox{green!30}{nikwax} ne tx direct \\
    3 & \colorbox{red!30}{dubman} brush & \colorbox{green!30}{denman} brush natural hair, \colorbox{green!30}{denman} brush, \colorbox{green!30}{denman} nylon brush, \colorbox{green!30}{denman} brush & \colorbox{green!30}{denman} brush & \colorbox{green!30}{denman} brush \\
    4 & \colorbox{red!30}{salamin} boots & mens \colorbox{green!30}{salomon} boots, \colorbox{green!30}{salomon} snowboard boots, \colorbox{green!30}{salomon} womens winter boots & \colorbox{green!30}{salomon} boots & \colorbox{green!30}{salomon} boots \\
    5 & \colorbox{red!30}{tumeric} soap bars & bali soap bars, camay soap bars, dettol soap bars, himalaya soap bars & \colorbox{green!30}{turmeric} soap bars & \colorbox{green!30}{turmeric} soap bars \\
    6 & doom eternal \colorbox{red!30}{puns} & doom eternal juguete,doom,doom eternal & doom eternal \colorbox{red!30}{puns} & doom eternal \colorbox{green!30}{pins} \\
    7 & \colorbox{red!30}{tonkatsu} ramen noodles & \colorbox{red!30}{tonkatsu} ramen noodles, \colorbox{red!30}{tonkatsu} ramen bowl, buldak ramen noodles, immi ramen noodles & \colorbox{green!30}{tonkotsu} ramen noodles & \colorbox{green!30}{tonkotsu} ramen noodles \\
    8 & \colorbox{red!30}{correlle} & \colorbox{red!30}{correlle} cafe red bowl,\colorbox{red!30}{correlle} plates white,\colorbox{red!30}{correlle},\colorbox{red!30}{correlle} red bowl & \colorbox{red!30}{correlle} & \colorbox{green!30}{corelle} \\
    9 & \colorbox{red!30}{laroche b5} & guy \colorbox{green!30}{laroche},flydigi b5,\colorbox{green!30}{laroche},\colorbox{green!30}{laroche} set & \colorbox{red!30}{laroche b5} & \colorbox{green!30}{la roche b5} \\
    \bottomrule
  \end{tabular}
  }
  \caption{Qualitative evaluation of RAG-generated spelling corrections across various examples. In 1--4, correctly spelled retrieved items lead to accurate corrections. In 5--6, the retrieved items do not involve misspelled spans of the input query, leading RAG generation to rely on the LLM's internal knowledge. In 7, the LLM generates the correct spelling despite misspelled retrieved items, while in 8, the model is misled by the incorrect retrieval. Finally, in 9, ``la roche'' and ``laroche'' are both real brands. The retriever does not correctly consider the context ``b5'' to distinguish the brands (``b5'' is a specific item that is only associated with brand ``la roche'').}
  \label{tab:qualitative_examples}
\end{table*}

Table~\ref{tab:retriever_performance} provides our primary results. We consider Mistral-7B and Clause-3-sonnet as the base LLMs. For each LLM, we evaluate our three different retrieval models. We also explore the role of the size of the product catalog by considering two different pools of documents: one with 572K documents (132K unique brands) and one with 60K documents (29K unique brands).

The overall best-performing setting is with ColBERT indexing 572K documents and providing retrieval results to Claude-3-sonnet. This configuration results in 39.3 F1 vs.~34.7 for Clause-3-sonnet used without retrieval (Table~\ref{tab:generative_performance}): a 4.6 point increase. Strikingly, the next best system is one that uses Mistral-7B, again with ColBERT indexing the larger document collection. This configuration achieves 35.9 vs.~28.1 for Mistral-7B used without retrieval: a 7.8 point increase. 

Table~\ref{tab:qualitative_examples} provides a more qualitative analysis that reveals nuanced interactions between retrieved context and LLM responses. When relevant context is retrieved (Examples 1--4), the LLM provides accurate corrections aligned with expected outputs. Conversely, in instances where context is absent (Examples 5--6), incorrect (Examples 7--8), or misleading (Example 9), the LLM's performance varied, highlighting the complex balance between the LLM's parameterized knowledge and the information retrieved. These examples illustrate that, while retrievers enrich context, they can also introduce noise or irrelevant data that might detract from accuracy if not carefully managed.

\begin{table*}[tp]
\centering
  \begin{tabular}{l l c c c}
    \toprule
    Retriever & LLM & Precision & Recall & F1 \\
    \midrule
    \multirow{2}{*}{None} 
    & Mistral-7B & 30.3 & 26.2 & 28.1 \\
    & Mistral-7B with Basic Fine-Tuning & 70.3 & 59.0 & 64.1 \\
    \midrule
    \multirow{3}{*}{Fuzzy BM25} 
    & Mistral-7B & 42.8 & 29.0 & 34.6\\
    & Mistral-7B with Basic Fine-Tuning & 49.7 & 40.3 & 44.5\\
    & Mistral-7B with Contextual Fine-Tuning & 77.4 & 59.5 & 67.3\\
    \midrule
    \multirow{3}{*}{ColBERT} 
    & Mistral-7B & 43.1 & 30.8 & 35.9\\
    & Mistral-7B with Basic Fine-Tuning & 52.3 & 42.1 & 46.6\\
    & Mistral-7B with Contextual Fine-Tuning & 77.6 & 65.5 & 71.0\\
    \bottomrule
  \end{tabular}
  \caption{Performance comparison across different retrieval configurations and fine-tuning setups. All experiments used an indexed pool of 572K documents (132K unique brands). The highest F1 score of 71.0 was achieved with ColBERT in RAG, demonstrating the added benefit of Contextual Fine-Tuning. These results indicate that fine-tuning with context-specific instructions is extremely effective.}
  \label{tab:overall_performance}
\end{table*}

\subsection{LLM Fine-Tuning}\label{sec:finetuned_rag}

Table~\ref{tab:overall_performance} summarizes our experiments involving LLM fine-tuning, using both Basic and Contextual variants of this method (Section~\ref{sec:finetune_data_elicit}). For these experiments, we adopt Mistral-7B as our base LLM. To facilitate comparisons, the top row in the first section of the table is repeated from Table~\ref{tab:generative_performance}, and the top rows from the middle and bottom sections are repeated from Table~\ref{tab:retriever_performance}. We include the precision/recall breakdown here to support further analysis of the trade-offs.

Across all three panels we see substantial gains from fine-tuning, with similar precision/recall ratios across all settings. The largest gains come from Contextual Fine-Tuning. The best performing configuration uses ColBERT and Contextual Fine-Tuning, leading to 70.1 F1, a 34.2 point increase over the system that employs only RAG with ColBERT. Thus, the overall message is very clear: if it is feasible to fine-tune the LLM with context, that is likely to lead to very substantial performance improvements.

Our primary goal is to improve performance on brands that are novel from the perspective of the LLM. Table~\ref{tab:performance_breakdown} seeks to quantify the extent to which Contextual Fine-Tuning marks progress in this area, as compared to Basic Fine-Tuning. In both cases, we train on the same set of examples (Section~\ref{sec:finetune_data_elicit}). The table shows that Contextual Fine-Tuning leads to a 6.9 point increase in overall F1 and a 16.7 point increase in queries containing brands. While brands remain very challenging, our approach certainly alleviates the challenge. 

These gains are supported by qualitative analysis as well. For example, in cases like ``snowflake necklace for women'', where the context includes varied necklaces (``swarovski snowflake necklace for women, efytal necklace for women, birthstone necklace for women, baguette necklace for women pavori''), Contextual Fine-Tuning helps the model produce the correct label ``swarovski snowflake necklace for women,'' ensuring that it makes thoughtful corrections rather than echoing the context.

\begin{table}
  \resizebox{1\linewidth}{!}{
  \setlength{\tabcolsep}{2pt}
  \begin{tabular}{@{} l c c @{}}
    \toprule
      & \multicolumn{2}{c}{F1} \\
      & All queries & Brands \\
    \midrule
    Basic Fine-Tuning & 64.1 & 44.1 \\
    RAG + Contextual Fine-Tuning & 71.0 & 60.8 \\
    \bottomrule
  \end{tabular}
  }
  \caption{Performance improvements from Contextual Fine-Tuning and RAG as compared to Basic Fine-Tuning without RAG, broken down by overall performance and performance on queries containing brands. The LLM is Mistral-7B and the retriever used for RAG is ColBERT over the 572K document collection.
  }
  \label{tab:performance_breakdown}
\end{table}

Ideally, we would be able to home in on a set of definitely new brands and measure performance on them. However, this set is challenging to define, since multiple sources of knowledge are in play, including the pretrained knowledge parameterized in LLM. However, we are able to identify a set of 525 brands that are in our evaluation set but absent from our fine-tuning dataset. For this set, we actually get an F1 score of 78.4 using Contextual Fine-Tuning and a ColBERT retriever. For future work, we will snapshot new brands by LLM release date as the cutoff, to present a more controlled experiment.

\subsection{Latency Considerations}\label{sec:latency}

Incorporating RAG leads to a slight latency increase; however, it remains within acceptable limits for real-time applications. Using the Mistral-7B model as a baseline, retrieval from a pool of 60K candidate documents adds only 2.42\% to the overall generation time, while expanding the pool to 572K documents results in a 2.79\% increase. These changes are minimal, and they enable substantial gains in accuracy.

\section{Conclusion}\label{sec:conclude}

In this paper, we introduced a fine-tuned Retrieval Augmented Generation (RAG) framework tailored for e-commerce spelling correction, specifically addressing the complexities posed by brand names and other non-standard lexicons. We showed that this approach is highly effective even with a frozen retriever and frozen large language model (Table~\ref{tab:retriever_performance}). In addition, we showed that fine-tuning the LLM with retrieved context leads to even larger gains (Table~\ref{tab:overall_performance}), particularly for spelling corrections involving evolving brand names (Table~\ref{tab:performance_breakdown}). These results underscore the value of incorporating retrieval and allowing the model to dynamically adapt to context in a way that standalone LLMs or RAG with frozen components cannot achieve.

Our qualitative analysis further revealed challenges inherent in using real-world data, such as the presence of misspellings in indexed documents, which can mislead the LLM during generation (Table~\ref{tab:qualitative_examples}). This highlights the importance of ensuring the quality of retrieved contexts. Practical improvements include refining the contextual data through heuristic signals, like user interactions and engagement metrics, to enhance relevance and accuracy. Another promising avenue is to diversify the styles and noise levels within the retrieved context to bolster the model’s robustness.

For long-term directions, we propose exploring mechanisms that enable LLMs to assess and selectively integrate context based on relevance and quality. Such advancements could pave the way for smarter, more context-aware LLMs that distinguish valuable insights from noise, ultimately enhancing their adaptability in real-world applications. Additionally, evaluating models with context on emerging entities could provide a more dynamic measure of RAG's effectiveness as new content enters the dataset. These lines of inquiry contribute insights into optimizing LLMs within RAG frameworks, driving advancements in the broader field of adaptive language models and their application in context-sensitive domains.

\section{Ethics Statements}

This study uses anonymized, user-generated data to enhance the model's ability to do contextual spelling correction in e-commerce. We acknowledge that user-generated data may reflect inherent biases, such as regional or demographic linguistic preferences, which could affect spelling correction accuracy for certain user groups. We are committed to monitoring these issues and improving the fairness of the model over time, aiming to make spelling correction equitable, inclusive, and accurate. Future efforts will focus on refining our methodology to address these concerns, especially as the model encounters new data and evolves to handle a broader range of brand-specific terminology and user inputs.

\bibliography{main}

\end{document}